\PassOptionsToPackage{unicode}{hyperref}
\PassOptionsToPackage{hyphens}{url}
\PassOptionsToPackage{dvipsnames,svgnames,x11names}{xcolor}
\documentclass[
  12pt]{article}

\usepackage{amsmath,amssymb}
\usepackage{enumitem}
\usepackage{xspace}
\usepackage{bm}
\usepackage{amsthm}
\usepackage{mathtools}
\usepackage{iftex}
\ifPDFTeX
  \usepackage[T1]{fontenc}
  \usepackage[utf8]{inputenc}
  \usepackage{textcomp} 
\else 
  \usepackage{unicode-math}
  \renewcommand{\bm}[1]{\symbf{#1}}
  \defaultfontfeatures{Scale=MatchLowercase}
  \defaultfontfeatures[\rmfamily]{Ligatures=TeX,Scale=1}
\fi
\usepackage{lmodern}
\ifPDFTeX\else  
\fi
\IfFileExists{upquote.sty}{\usepackage{upquote}}{}
\IfFileExists{microtype.sty}{
  \usepackage[]{microtype}
  \UseMicrotypeSet[protrusion]{basicmath} 
}{}
\makeatletter
\@ifundefined{KOMAClassName}{
  \IfFileExists{parskip.sty}{%
    \usepackage{parskip}
  }{
    \setlength{\parindent}{0pt}
    \setlength{\parskip}{6pt plus 2pt minus 1pt}}
}{
  \KOMAoptions{parskip=half}}
\makeatother
\usepackage{xcolor}
\makeatletter
\ifx\paragraph\undefined\else
  \let\oldparagraph\paragraph
  \renewcommand{\paragraph}{
    \@ifstar
      \xxxParagraphStar
      \xxxParagraphNoStar
  }
  \newcommand{\xxxParagraphStar}[1]{\oldparagraph*{#1}\mbox{}}
  \newcommand{\xxxParagraphNoStar}[1]{\oldparagraph{#1}\mbox{}}
\fi
\ifx\subparagraph\undefined\else
  \let\oldsubparagraph\subparagraph
  \renewcommand{\subparagraph}{
    \@ifstar
      \xxxSubParagraphStar
      \xxxSubParagraphNoStar
  }
  \newcommand{\xxxSubParagraphStar}[1]{\oldsubparagraph*{#1}\mbox{}}
  \newcommand{\xxxSubParagraphNoStar}[1]{\oldsubparagraph{#1}\mbox{}}
\fi
\makeatother

\usepackage{longtable,booktabs,array}
\usepackage{calc} 
\usepackage{etoolbox}
\makeatletter
\patchcmd\longtable{\par}{\if@noskipsec\mbox{}\fi\par}{}{}
\makeatother
\IfFileExists{footnotehyper.sty}{\usepackage{footnotehyper}}{\usepackage{footnote}}
\makesavenoteenv{longtable}
\usepackage{graphicx}
\makeatletter
\def\maxwidth{\ifdim\Gin@nat@width>\linewidth\linewidth\else\Gin@nat@width\fi}
\def\maxheight{\ifdim\Gin@nat@height>\textheight\textheight\else\Gin@nat@height\fi}
\makeatother
\setkeys{Gin}{width=\maxwidth,height=\maxheight,keepaspectratio}
\makeatletter
\def\fps@figure{htbp}
\makeatother

\makeatletter
\@ifpackageloaded{caption}{}{\usepackage{caption}}
\AtBeginDocument{%
\ifdefined\contentsname
  \renewcommand*\contentsname{Table of contents}
\else
  \newcommand\contentsname{Table of contents}
\fi
\ifdefined\listfigurename
  \renewcommand*\listfigurename{List of Figures}
\else
  \newcommand\listfigurename{List of Figures}
\fi
\ifdefined\listtablename
  \renewcommand*\listtablename{List of Tables}
\else
  \newcommand\listtablename{List of Tables}
\fi
\ifdefined\figurename
  \renewcommand*\figurename{Figure}
\else
  \newcommand\figurename{Figure}
\fi
\ifdefined\tablename
  \renewcommand*\tablename{Table}
\else
  \newcommand\tablename{Table}
\fi
}
\@ifpackageloaded{float}{}{\usepackage{float}}
\floatstyle{ruled}
\@ifundefined{c@chapter}{\newfloat{codelisting}{h}{lop}}{\newfloat{codelisting}{h}{lop}[chapter]}
\floatname{codelisting}{Listing}

\makeatother
\makeatletter
\@ifpackageloaded{caption}{}{\usepackage{caption}}
\@ifpackageloaded{subcaption}{}{\usepackage{subcaption}}
\makeatother

\ifLuaTeX
  \usepackage{selnolig}  
\fi
\usepackage[]{natbib}
\usepackage{bookmark}

\IfFileExists{xurl.sty}{\usepackage{xurl}}{} 
\usepackage{algorithm}
\usepackage{algpseudocode}
\algrenewcommand\algorithmicrequire{\textbf{Input:}}
\algrenewcommand\algorithmicensure{\textbf{Output:}}
\hypersetup{
  pdftitle={Amortized Nonlinear Variable Selection with Prior-Data Fitted Networks},
  pdfauthor={Jin Zhu; Junxian Zhu},
  pdfkeywords={Variable selection, Markov blanket, Transformer, Prior-data fitted networks},
  colorlinks=true,
  linkcolor={blue},
  filecolor={Maroon},
  citecolor={Blue},
  urlcolor={Blue},
  pdfcreator={LaTeX via pandoc}}

\newtheorem{theorem}{Theorem}
\newtheorem{lemma}[theorem]{Lemma}
\newtheorem{proposition}[theorem]{Proposition}

\newtheorem{assumption}{Assumption}

\newcommand{\anon}{1}

\newcommand{\sig}{\operatorname{sigmoid}}

\begin{document}

\def\spacingset#1{\renewcommand{\baselinestretch}%
{#1}\small\normalsize} \spacingset{1}


\if1\anon
{
  \title{\bf Amortized Nonlinear Variable Selection with Prior-Data Fitted Networks}
  \author{Jin Zhu and Junxian Zhu\thanks{Jin Zhu and Junxian Zhu contributed equally to this work and are listed in alphabetical order.} \hspace{.2cm}\\
    University of Birmingham, \\
    University of Science and Technology of China}
  
  \date{}
  \maketitle
} \fi

\if0\anon
{
  \bigskip
  \bigskip
  \bigskip
  \begin{center}
    {\LARGE\bf Title}
\end{center}
  \medskip
} \fi

\bigskip
\begin{abstract}
Variable selection is an important task for interpreting potentially complex systems in fields such as biology, business, and sociology. The importance of variable selection has attracted considerable attention in statistics and machine learning. In this paper, we are mainly interested in model-agnostic variable selection for identifying variables that have general nonlinear dependence on the response. We propose MambaPFN, short for \underline{m}odel-\underline{a}gnostic \underline{M}arkov \underline{b}l\underline{a}nket selection with \underline{p}rior-data \underline{f}itted \underline{n}etworks, which is built upon the Transformer, a building block of many popular large language models. Specifically, we consider a functional-mapping perspective on variable-selection learning, in which the Transformer is treated as a learnable mapping from a dataset to its relevant variables. To learn this mapping, MambaPFN introduces three new components: (i) a faithful, rich, and scalable scheme for generating pretraining data, (ii) an objective that targets each variable's posterior inclusion probability, and (iii) a modified Transformer architecture tailored to variable selection. Once pretrained, MambaPFN is applied to new datasets without parameter updates. Empirically, MambaPFN achieves desirable variable-selection performance in simulation studies and exhibits a scaling law---more pretraining datasets and larger models can continuously improve its performance. Applied to a Facebook comments dataset, MambaPFN selects the fewest variables and achieves the highest computational efficiency and predictive performance, improving upon the best baselines by about 20\%.
\end{abstract}

\noindent%
{\it Keywords:} Variable selection, Markov blanket, Transformer, Prior-data fitted networks
\vfill

\newpage
\spacingset{1.8} 

\section{Introduction}

Variable selection aims to identify the most relevant and informative variables in a dataset while removing irrelevant, noisy, or redundant ones. It is central to many scientific and societal applications because it can provide insight into the underlying system. For example, in biomedicine, selecting informative variables from high-dimensional omics and single-cell data supports reliable biomarker discovery \citep{hedou2024stabl}. In materials science, selecting compact physicochemical descriptors can reveal structure--property relationships and guide catalyst design \citep{han2024electrocatalytic}. Furthermore, variable selection can improve robustness and generalization while reducing effective model complexity \citep{li2024universum,liao2024sparse}. Therefore, variable selection remains a fundamental and active problem in machine learning and data-driven science. 

\textbf{Literature review}. Because of the importance of variable selection, dozens of methods have been developed over the last few decades; we review representative ones here. Existing variable selection methods can be divided mainly into two categories: \underline{model-based} and \underline{model-agnostic} methods.

\textit{Model-based variable selection} imposes explicit structure on the regression function. For linear models, sparsity is induced through convex or non-concave penalties \citep{tibshirani1996regression,fan2001variable,zhang2010nearly,shen2012likelihood} or iterative algorithms \citep{blumensath2009iterative,huang2018constructive,zhu2020polynomial}. These methods have been extended to more complicated models, such as models in which predictors have a known group structure \citep{yuan2006model}, models in which hierarchical penalties enforce heredity among main and interaction effects \citep{bien2013lasso,lim2015learning}, single-index models \citep{kong2007variable}, and sparse additive models \citep{ravikumar2009sparse,meier2009high,huang2010variable}. These methods yield interpretable estimators and, under suitable conditions, strong selection guarantees. Their targets and guarantees, however, are tied to the assumed structure.

\textit{Model-agnostic variable selection} is typically implemented via statistics-based methods and machine-learning (ML)-based methods.
\begin{itemize}[leftmargin=*]
\item \textit{Statistics-based methods}. Notable examples are knockoff procedures, which construct negative-control variables and use them to control false discoveries in variable selection \citep{candes2018panning,romano2020deep,bates2021metropolized,zhu2021deeplink,spector2022powerful,ren2023derandomizing}. Kernel-based methods instead identify variables through gradients of conditional dependence measures in reproducing kernel Hilbert spaces or through conditional covariance operators \citep{yamada2014high,yang2016model,chen2017kernel,climente2019block,poignard2020sparse}.
\item \textit{ML-based methods} select relevant variables using ML algorithms such as neural networks. For example, because neural networks have a strong capacity to approximate arbitrary nonlinear relationships, various methods have been developed to select variables that have complicated relationships with the response \citep{yoon2019invase,yamada2020feature,lemhadri2021lassonet,chen2021nonlinear,luo2023sparse,jiang2024protogate,yang2024enns}.
\end{itemize}

\subsection{Motivation: The Bitter Lesson}

\citet{sutton2019bitter} presents the view that, in the long run, general approaches that scale with available computational power tend to outperform those based on domain-specific understanding because they are better able to take advantage of the falling cost of computation over time. This principle was proposed in 2019 and has gained increasing acceptance as the capacity of large language models has grown, e.g., ChatGPT \citep{openai2026gpt6}, Claude \citep{anthropic2026fable}, and DeepSeek \citep{deepseekai2026deepseekv4}, enabling them to solve increasingly challenging problems.

However, none of the aforementioned methods implements this viewpoint because they are not amortized across datasets; in other words, given a new dataset, they need to refit a model, whether it is a neural network or a linear model, to perform variable selection. Furthermore, non-amortized methods are static and cannot exhibit a scaling-law effect across datasets, under which the performance of an amortized method would reasonably improve if more diverse datasets were used for training. At the same time, neural-network-based variable-selection methods, particularly those without distributional assumptions on the data, can be time-consuming \citep[e.g.,][]{chen2021nonlinear}. From the perspective of model deployment, because the variable-selection step aims to reduce the variable set before fitting a model, a time-consuming selection procedure is less desirable in applications.

On the other hand, methods that adopt this viewpoint are achieving increasingly strong performance through amortized learning. For example, prior-data fitted networks (PFNs) apply this amortized-learning paradigm: a model is pretrained once and then makes predictions on new datasets without retraining \citep{muller2022transformers}. Subsequent methods have improved their predictive accuracy, scalability, and coverage of real-data tasks \citep{hollmann2023tabpfn,grinsztajn2025tabpfn,zhang2025limix,ma2026tabdpt,qu2026tabiclv2}, together with a complementary theoretical understanding of these methods \citep{ma2025optimal}. Amortization has also moved beyond point prediction to tasks such as causal-structure identification \citep{lorch2022amortized,peng2026causcale,thompson2026arrow}, (conditional) average treatment effect estimation \citep{balazadeh2026causalpfn,robertson2025dopfn,sauter2025activa}, interval estimation \citep{ye2026amotized}, and Bayesian optimization \citep{muller2023pfns4bo,rakotoarison2026alphapfn}.

These advances raise a natural question: can variable selection itself be amortized, so that the relevant variables of a new dataset are obtained without any task-specific optimization? Existing amortized models do not answer this question directly: PFN-based predictors do not output selected variables, and amortized causal-discovery methods must first recover the whole causal graph, a harder task than identifying the Markov blanket of the response.

\subsection{Proposal and contributions}
In this paper, we introduce a variable-selection method for identifying the minimal Markov blanket of the response without prespecifying the form of the regression function. The method is amortized: a Transformer is pretrained on a large collection of variable-selection tasks and can then be applied to a new dataset without parameter updates, yielding substantial computational savings.

The paper makes two main contributions.
\begin{itemize}[leftmargin=*]
  \item Methodologically, we introduce three key components for learning a suitable Transformer for amortized variable selection: (i) a faithful, rich, and scalable scheme for generating pretraining data; (ii) an objective that guides the Transformer toward estimating each variable's posterior inclusion probability; and (iii) a modified Transformer architecture tailored to variable selection.
  \item Empirically, the proposal achieves higher nonlinear variable-selection accuracy than representative model-agnostic methods in our simulation studies. We also observe empirical scaling behavior: the learned Transformer performs better as either the diversity of the pretraining datasets or the Transformer model size increases. The analysis of the Facebook comments data provides further empirical evidence of the proposal's practical utility: it selects the fewest variables while achieving the highest predictive performance. It also performs variable selection efficiently, providing more than a 10-fold speedup over the neural-network-based baseline while remaining computationally competitive with linear-model-based methods such as Lasso.
\end{itemize}

\subsection{Organization}

The remainder of this paper is organized as follows. In Section~\ref{sec:preliminaries}, we introduce some basic concepts and techniques used in this paper. Section~\ref{sec:method} introduces the proposed method. Sections~\ref{sec:simulation} and~\ref{sec:real-data} present numerical analyses of synthetic datasets and a real-world dataset, respectively. We conclude the paper and provide a discussion in Section~\ref{sec:conclusion}. Technical proofs and experimental details are deferred to the Appendix.

\section{Preliminaries}\label{sec:preliminaries}
\subsection{Problem formulation}
This paper concerns variable selection in a general regression setting. Let $Y \in \mathbb{R}$ be the response, i.e., the random variable of interest, and let \(X=(X_1,\ldots,X_p)^\top\in\mathbb{R}^p\) be a $p$-dimensional vector of covariates used to model the response. Throughout the paper, we impose the following assumption. 
\begin{assumption}\label{con:positive-density}
The probability density function of $(X_1,\ldots,X_p, Y)$ is strictly positive on $\mathbb{R}^{p+1}$. 
\end{assumption}
This assumption can be interpreted as an identifiability condition; for example, it excludes cases in which two variables follow an exact deterministic relationship (e.g., a linear relationship). Similar identifiability conditions are imposed in the literature to avoid such ill-posed cases \citep{zhang2010nearly}. Moreover, this assumption is weak overall because it does not rely on any model structure, such as the linear model or the generalized additive model. This makes the assumption well suited to our goal of selecting variables with nonlinear effects. 

Under Assumption~\ref{con:positive-density}, there exists a unique minimal set $S^\star \subseteq[p]$ \citep{dawid1979conditional,pearl2014probabilistic} such that the sub-vector $X_{S^\star}$ of $X$ indexed by $S^\star$ satisfies
\[
Y \perp X_{[p]\setminus S^\star} \mid X_{S^\star},
\]
This means that, given $X_{S^\star}$, the remaining variables $X_{[p]\setminus S^\star}$ are independent of $Y$; in other words, $X_{S^\star}$ is the smallest set of variables sufficient to explain $Y$. Thus, $X_{S^\star}$, which we call the Markov blanket of $Y$, is the target of this paper.\footnote{It is also called the Markov boundary in some references; throughout the paper, we use the term Markov blanket.}

In practice, $S^\star$ is unknown, and we typically only have a dataset $\mathcal{D}_n=\{(X_i,Y_i)\}_{i=1}^n$ where $(X_i,Y_i) \in \mathbb{R}^{p+1}$ are independently and identically drawn from the joint distribution of $(X, Y)$. Our goal is to identify $S^\star$ from $\mathcal{D}_n$. 

\subsection{Transformer}\label{sec:transformer}

The core module of our amortized learner is the Transformer, which was first introduced by \citet{vaswani2017attention} and is widely used in modern large language models. A notable property of the Transformer is its ability to make predictions from variable-length input sequences by leveraging interactions among the sequence elements. It achieves this property through several key components: embeddings, self-attention, and multilayer perceptrons, which are described below.

\textbf{Embedding}. A Transformer first maps its input tokens to a common representation space. Given $n$ scalars $x_1,\ldots,x_n$, a learnable embedding matrix $\bm{E}$ produces
\begin{equation}
\bm{H}=(\bm{E} x_1,\ldots, \bm{E} x_n)^\top
\in\mathbb{R}^{n\times m}.
\label{eq:embedding}
\end{equation}
\textbf{Self-attention module}. Given a token matrix $\bm{H}\in\mathbb{R}^{n\times m}$, self-attention forms query, key, and value matrices $\bm{H}\bm{W}_Q$, $\bm{H}\bm{W}_K$, and $\bm{H}\bm{W}_V$, where $\bm{W}_Q,\bm{W}_K,\bm{W}_V\in\mathbb{R}^{m\times d}$ are learnable parameters, and aggregates the value representations using normalized query--key similarities. With an additional learnable $\bm{W}_O\in\mathbb{R}^{d\times m}$ and an identity mapping, the self-attention map is
\begin{equation}
\texttt{SA}(\bm{H})
=\bm{H}
+\texttt{Softmax}\!\left(
\frac{(\bm{H}\bm{W}_Q)(\bm{H}\bm{W}_K)^\top}{\sqrt d}
\right)(\bm{H}\bm{W}_V)\bm{W}_O.
\label{eq:self-attention}
\end{equation}
For simplicity, we omit the positional embedding in Equation~\eqref{eq:embedding}, multi-head attention and layer normalization from Equation~\eqref{eq:self-attention}, as in \citet{kim2023provable, kajitsuka2024transformers}. 

\textbf{Multilayer perceptron}. The multilayer perceptron (MLP) applies the same two-layer mapping independently to each row of an input $\bm{Z}\in\mathbb{R}^{n\times m}$. Let $h$ be the hidden width. The MLP introduces the learnable weight matrices $\bm{W}_1\in\mathbb{R}^{m\times h}$ and $\bm{W}_2\in\mathbb{R}^{h\times m}$ and the learnable bias vectors $\bm{b}_1\in\mathbb{R}^{h}$ and $\bm{b}_2\in\mathbb{R}^{m}$ to process $\bm{Z}$ via the following rule:
\begin{align*}
\texttt{MLP}(\bm{Z})
= \bm{Z}
+ \sigma\!\left(\bm{Z}\bm{W}_1+\bm{1}_n\bm{b}_1^\top\right)\bm{W}_2
+ \bm{1}_n\bm{b}_2^\top \in \mathbb{R}^{n\times m},
\end{align*}
where $\bm{1}_n$ is the $n$-dimensional vector of ones and $\sigma$ is an element-wise activation function. 

The complete Transformer alternately stacks multiple \texttt{SA} and \texttt{MLP} modules and takes the form
\begin{align*}
n^{-1}\textbf{1}_n^\top (\texttt{MLP}_K \circ \texttt{SA}_K\circ\cdots\circ\texttt{MLP}_1 \circ \texttt{SA}_1 (\bm{H})) \bm{w}_U,
\end{align*}
where $n^{-1}\textbf{1}^\top_n$ averages the representations over all samples, and $\bm{w}_U \in \mathbb{R}^{m}$ is a learnable parameter that produces a scalar output. 

\section{Methodology}\label{sec:method}

\subsection{Variable selection: a mapping perspective} 
Variable selection can be interpreted from a mapping perspective. Specifically, $S^\star$ can be equivalently represented as a $p$-dimensional indicator vector $I^* = (I_1, \ldots, I_p) \in \{0, 1\}^{p}$, where $I_j = \mathbb{I}(j \in S^\star)$. Variable selection seeks to obtain a function $T$ that maps a dataset $\mathcal{D}_n$ to an indicator vector $\widehat{I}$ that indicates which variables should be selected. This perspective can also be used to interpret existing variable selection methods, such as the Lasso. Given a dataset, the Lasso learns a sparse coefficient vector under the linear model assumption $\mathbb{E}[Y|X] = X^\top \beta$ and then derives a variable selection result $\widehat{I}$ from the estimated coefficient $\widehat{\beta}$.

In this section, we introduce a learning-based method for obtaining a function $T$ that can perform variable selection. Once $T$ is learned, it can be applied directly to new datasets to select variables. Learning such a function raises three questions:
\begin{itemize}[leftmargin=*]
\item The first question is how to collect pairs of datasets and their true variable indicators for learning an effective function $T$. This question arises because the true variable indicators are typically unknown in practice.
\item The second question is how to design an objective function so that the learned function $T$ outputs an $\widehat{I}$ that is close to $I^*$.
\item The last question is how to design the function $T$ itself. For example, it should accommodate various input shapes, as the input datasets can have different $n$ and $p$. 
\end{itemize}
We answer these three questions in Sections~\ref{sec:method-data}, \ref{sec:opt-objective}, and \ref{sec:transformer-vs}, respectively.

\subsection{Pretraining data generation}\label{sec:method-data}

In this section, we describe our approach to generating the pretraining data $(\mathcal{D}_n, I^*)$. Ideally, the generation procedure should satisfy three principles:
\begin{itemize}[leftmargin=*]
  \item \textit{Computational scalability}: generating $(\mathcal{D}_n, I^*)$ should be computationally convenient so that we can generate a large amount of pretraining data and increase the likelihood of successfully training the mapping. 
  \item \textit{Distributional richness}: the generated probability density functions should represent a wide range of relationships between the covariates and the response, thereby supporting the goal of nonlinear variable selection. 
  \item \textit{Faithfulness}: the probability density function used to generate the dataset $\mathcal{D}_n$ should indeed correspond to $I^*$ and should also satisfy Assumption~\ref{con:positive-density}. 
\end{itemize}

\begin{algorithm}[!htbp]
\spacingset{1}
\caption{Generation of a pretraining pair $(\mathcal{D}_n, I^{\psi})$}\label{alg:data-generation}
\begin{algorithmic}[1]
\Require Graph class $\mathcal{G}$; function class $\mathcal{F}$; family $\mathcal{P}^{(0)}$ of strictly positive densities.
\Ensure dataset $\mathcal{D}_n=\{(X_i,Y_i)\}_{i=1}^n$ and indicator vector $I^{\psi}=(I^{\psi}_1,\ldots,I^{\psi}_p)\in\{0,1\}^p$.
\State Draw a causal graph $\psi$ from $\mathcal{G}$ on the nodes $V_1=X_1,\ldots,V_p=X_p$ and $V_{p+1}=Y$.
\State For $j=1,\ldots,p$, $I^{\psi}_j=1$ if $X_j$ is a parent, child, or spouse of $Y$; otherwise $I^{\psi}_j=0$.
\State Partition the nodes of $\psi$ into topological layers $\mathcal{L}_0,\ldots,\mathcal{L}_K$.
\For{$j\in\mathcal{L}_0$} \Comment{root nodes}
  \State Draw $p^{(0)}_j$ from $\mathcal{P}^{(0)}$ and sample $V_{1j},\ldots,V_{nj}\overset{\text{i.i.d.}}{\sim}p^{(0)}_j$.
\EndFor
\For{$k=1,\ldots,K$} \Comment{layer-wise sampling}
  \For{$j\in\mathcal{L}_k$}
    \State Draw $f_j$ from $\mathcal{F}$ and a noise distribution $q_j$ from $\mathcal{P}^{(0)}$.
    \State For $i=1,\ldots,n$, set $V_{ij}=f_j\big(V_{i,\operatorname{pa}_{\psi}(j)}\big)+\varepsilon_{ij}$ where $\varepsilon_{ij}\sim q_j$.
  \EndFor
\EndFor
\State Set $X_i=(V_{i1},\ldots,V_{ip})^\top$ and $Y_i=V_{i,p+1}$ for $i=1,\ldots,n$.
\end{algorithmic}
\end{algorithm}

To this end, we introduce a graph-based method, whose complete procedure is summarized in Algorithm~\ref{alg:data-generation}. Specifically, we start with a directed acyclic graph $\psi$ randomly drawn from $\mathcal{G}$, a class of directed acyclic graphs on $p+1$ nodes. The graph $\psi$ characterizes the causal relationships among $(X_1, \ldots, X_p, Y)$, where each node in the graph corresponds to one variable. Since $\psi$ is known, we can directly extract the Markov blanket of $Y$ and thereby find $I^{\psi}$. Specifically, the Markov blanket of $Y$ is the union of (i) the parents of \(Y\), (ii) the children of \(Y\), and (iii) the other parents of those children.

Given the graph $\psi$, we then generate a probability distribution. The density associated with $\psi$ admits a factorization. Specifically, let $V_i = X_i$ denote the $i$th node in $\psi$ for $i=1, \ldots, p$, and let $V_{p+1}=Y$ denote the last node. The nodes of $\psi$ can then be partitioned into topological layers $\mathcal{L}_0,\ldots,\mathcal{L}_K$, where $\mathcal{L}_0=\{j:\operatorname{pa}_{\psi}(j)=\varnothing\}$ contains the root nodes, and, for $k\geq 1$, every parent of a node in $\mathcal{L}_k$ belongs to an earlier layer. The resulting joint density therefore factorizes as
\begin{equation}\label{eq:prob-factorize}
\prod_{j\in\mathcal{L}_0} p^{(0)}_j(v_j)
\prod_{j\in\mathcal{L}_1} p^{(1)}_j\!\left(v_j\mid v_{\operatorname{pa}_{\psi}(j)}\right) 
\times \cdots \times
\prod_{j\in\mathcal{L}_K} p^{(K)}_j\!\left(v_j\mid v_{\operatorname{pa}_{\psi}(j)}\right).
\end{equation}

Next, we specify the setup for the distribution $p^{(0)}$ and the conditional distributions $p^{(1)}, \ldots, p^{(K)}$. For the distribution $p^{(0)}$, to satisfy Assumption~\ref{con:positive-density}, we consider a set of common distributions $\mathcal{P}^{(0)}$, each of which has a strictly positive density. For the conditional distributions $p^{(k)}_j(v_j | v_{\operatorname{pa}_{\psi}(j)})$, we adopt the following structural model:
\begin{equation}\label{eq:structure-model}
  V_j=f\!\left(V_{\operatorname{pa}_\psi(j)}\right)+\varepsilon_j,
\end{equation}
where $\varepsilon_j$ is an independent random variable associated with a distribution in $\mathcal{P}^{(0)}$, and $f$ is a function randomly drawn from a predefined function class $\mathcal{F}$ comprising various functions. The function $f$ should take all parents of $V_j$ as arguments. We denote the resulting joint distribution by $P_{\psi, f}$ and the collection of all possible joint distributions by $\mathcal{P}$.

Next, we examine whether the above procedure satisfies computational scalability, distributional richness, and faithfulness. First, the dataset $\mathcal{D}_n$ is computationally convenient to construct. Equation~\eqref{eq:prob-factorize} naturally induces a layer-wise sampling algorithm. We first sample all root variables in $\mathcal{L}_0$. Then, for $k=1,\ldots,K$, once the variables in the preceding layers have been generated, we sample all variables in $\mathcal{L}_k$ according to Equation~\eqref{eq:structure-model}. A complete observation is obtained after a pass through all topological layers. Second, the family $\mathcal{P}$ is distributionally rich because the structural model allows various functional combinations and $\mathcal{G}$ can generate various graph structures. The following result illustrates this richness by showing that $\mathcal{P}$ includes common statistical models. 
\begin{proposition}\label{prop:coverage}
The structural model includes generalized additive models and interaction models for the response $Y$ as special cases. 
\end{proposition}
Finally, we examine faithfulness, $I^{\psi} = I^*$. Let $\textup{MB}_{\psi}(Y)$ denote the Markov blanket of $Y$ in $\psi$. Since the densities in $\mathcal{P}^{(0)}$ are strictly positive, the joint density~\eqref{eq:prob-factorize} is also strictly positive; hence, Assumption~\ref{con:positive-density} holds, which implies $S^\star \subseteq \textup{MB}_{\psi}(Y)$ \citep{koller2009probabilistic,pearl2014probabilistic}. The reverse inclusion may fail because the dependence of $Y$ on a covariate can cancel out. For example, if $Y=X_1+\varepsilon$ and $X_2=Y+X_1+\varepsilon'$ with $\varepsilon,\varepsilon'\sim N(0,1)$, then $X_1\in\textup{MB}_{\psi}(Y)$, yet $Y\perp X_1\mid X_2$ and thus $X_1\notin S^\star$. Another example concerns the specification of $\mathcal{F}$: if a child $Z$ of $Y$ is generated by $Z=Y\,\mathbb{I}(X_1>0)+X_2\,\mathbb{I}(X_1\le0)+\varepsilon$, then $Y, X_1, X_2$ all affect $Z$. Because $Y$ and $X_2$ never jointly affect $Z$, we have $Y\perp X_2\mid X_{1}, Z$, although $X_2\in\textup{MB}_{\psi}(Y)$.

Lemma~\ref{lemma:graph-design} shows how such cancellations can be avoided under the following restrictions. 
\begin{assumption}\label{con:generate-graph}
Let $\psi$ be a graph drawn from $\mathcal{G}$. No $X_j\in\textup{MB}_{\psi}(Y)$ satisfies any of the following conditions: (A) $X_j$ is a parent of both $Y$ and a child of $Y$; (B) $X_j$ is a child of $Y$ and a parent of another child of $Y$; (C) $X_j$ is a parent of at least two children of $Y$.
\end{assumption}
\begin{assumption}\label{con:generate-function}
  Let $f\in\mathcal{F}$ be a mapping from $\mathbb{R}^m$ to $\mathbb{R}$. For each $i\in\{1,\ldots,m\}$ and almost every $u^*_j$ for $j\neq i$, the function $u_i\mapsto f(u^*_1,\ldots,u^*_{i-1},u_i,u^*_{i+1},\ldots,u^*_m)$ is not almost everywhere constant.
\end{assumption}
\begin{assumption}\label{con:generate-noise}
  Every distribution in $\mathcal{P}^{(0)}$ has a log-concave density $g$, i.e., $g(\lambda s+(1-\lambda)t)\ge g(s)^{\lambda}g(t)^{1-\lambda}, \text{ for all } s,t\in\mathbb{R} \text{ and } \lambda\in[0,1].$
\end{assumption}
\begin{lemma}\label{lemma:graph-design}
For the joint density given by Equations~\eqref{eq:prob-factorize}--\eqref{eq:structure-model}, we have $\textup{MB}_{\psi}(Y)=S^\star$ if Assumptions~\ref{con:generate-graph}--\ref{con:generate-noise} hold. 
\end{lemma}
Lemma~\ref{lemma:graph-design} imposes assumptions on $\mathcal{G}$, $\mathcal{F}$, and $\mathcal{P}^{(0)}$ to rule out the examples above. Assumption~\ref{con:generate-graph} is not restrictive because it concerns only the \textit{local} structure of $\psi$ around $Y$. For example, it holds whenever $Y$ has no children, which already allows Proposition~\ref{prop:coverage} to hold, as shown in its proof. Assumption~\ref{con:generate-function} excludes scenarios in which two parents never act together. It can be further relaxed to a local condition on the functions around $Y$, analogous to Assumption~\ref{con:generate-graph}. Assumption~\ref{con:generate-noise} is satisfied by many common distributions, such as the Gaussian, Laplace, and logistic distributions; in particular, it covers the Gaussian noise frequently assumed in the variable-selection literature \citep{zhang2010nearly,huang2018constructive,wainwright2019high}. Assumption~\ref{con:generate-noise} can be further relaxed because the proof requires it only for the noise terms associated with the children of $Y$. More importantly, these assumptions are easy to implement in practice and ensure faithfulness. 

\subsection{Optimization objective}\label{sec:opt-objective}

We seek a dataset-level function $T_{\theta}$ mapping $\mathcal{D}_n$ to $\{0,1\}^p$ that minimizes the misclassification loss $\frac{1}{p}\sum_{j=1}^p \mathbb{I}([T_{\theta}(\mathcal{D}_n)]_j \neq I^{\psi}_j)$, where $\theta$ denotes the learnable parameter vector. An objective value of zero corresponds to exact variable selection. However, directly minimizing this objective is challenging as the indicator function is non-differentiable. We therefore consider the following differentiable elementwise Bernoulli log-loss
\begin{equation}
\mathcal{L}(\theta;\mathcal{D}_n,I^{\psi})
=-\frac{1}{p}\sum_{j=1}^{p}
\Big\{
I^{\psi}_j\log[T_\theta(\mathcal{D}_n)]_j
+(1-I^{\psi}_j)\log\big(1-[T_\theta(\mathcal{D}_n)]_j\big)
\Big\}. 
\label{eq:sel-loss}
\end{equation}
This objective function is easy to implement using standard deep-learning frameworks, which also provide various optimizers. 

Proposition~\ref{prop:objective} reveals the population-level target of the optimization problem in~\eqref{eq:sel-loss}. Specifically, let $\mathcal{R}(\theta) =\mathbb{E}_{\mathcal{D}_n\sim \mathcal{P}} \left[\mathcal{L}(\theta;\mathcal{D}_n,I^{\psi})\right] \coloneq \mathbb{E}_{P_{\psi, f}\sim \mathcal{P}}\, \mathbb{E}_{\mathcal{D}_n \sim P_{\psi,f}} \left[\mathcal{L}(\theta;\mathcal{D}_n,I^{\psi})\right]$ be the population loss. Here, we make two notational simplifications: (i) we use $P_{\psi, f} \sim \mathcal{P}$ to represent a distribution drawn from the family $\mathcal{P}$, and (ii) we omit the subscripts of expectations whenever this causes no confusion. For each $j\in \{1, \ldots, p\}$, let $\mathbb{P}_{\psi \sim \mathcal{P}}\left(I^{\psi}_j=1 \mid\mathcal{D}_n\right)$ be the posterior inclusion probability of the $j$th covariate given a finite-sample dataset $\mathcal{D}_n$. Note that this posterior depends only on the family $\mathcal{P}$ used in the pretraining data-generation process. Proposition~\ref{prop:objective} shows that if $T_{\theta}$ belongs to a sufficiently large function class, the global minimizer of the population risk identifies the posterior inclusion probabilities. 
\begin{proposition}[Posterior inclusion probabilities]\label{prop:objective}
  Under Assumptions~\ref{con:generate-graph}--\ref{con:generate-noise}, if there exists $\bar\theta$ such that $T_{\bar\theta,j}(\mathcal{D}_n)=\mathbb{P}_{\psi \sim \mathcal{P}}\left(I^{\psi}_j=1 \mid\mathcal{D}_n\right)$ for every $j$, then $\mathcal{R}(\theta) \geq \mathcal{R}(\bar\theta)$ for any $\theta$. 
\end{proposition}
Furthermore, when $\mathcal{D}_n$ contains sufficiently many observations, a stronger result holds. 
\begin{proposition}[Large-sample behavior]\label{prop:posterior-limit}
Under the same notation and assumptions as in Proposition~\ref{prop:objective}, $T_{\bar\theta,j}(\mathcal{D}_n) \to I^{\psi}_j$ as $n \to \infty$.
\end{proposition}
This proposition shows that the solution to the global optimization problem converges to the deterministic true label $I^{\psi}_j$. 

In practice, we often encounter finite-sample datasets; thus, for $\widehat{\theta}$ obtained by optimizing Equation~\eqref{eq:sel-loss}, $T_{\widehat\theta}(\mathcal{D}_n)$ lies in $[0, 1]^{p}$. To obtain the variable-selection results, we apply a hard threshold to the estimated inclusion probabilities $T_{\widehat\theta}(\mathcal{D}_n)$. Specifically, we use $\alpha\in(0,1)$ to obtain $\widehat{I}=(\widehat{I}_1, \ldots, \widehat{I}_p)$, where $\widehat I_j = \mathbb{I}\!\left\{[T_{\widehat\theta}(\mathcal{D}_n)]_j>\alpha\right\}$. Given a grid of candidate $\alpha_1, \ldots, \alpha_M$, the threshold $\alpha$ is selected by the following three steps:
\begin{enumerate}[leftmargin=*]
  \item We use the same procedure described in Section~\ref{sec:method-data} to generate pairs of $(\mathcal{D}_n, I^*)$. 
  \item Apply $T_{\widehat\theta}$ and each candidate $\alpha_m$ to $\mathcal{D}_n$ to obtain $\widehat{I}^{(\alpha_m)}$, and compute the mis-selection rate $\frac{1}{p}\sum_{j=1}^p \mathbb{I}\!\left\{\widehat{I}^{(\alpha_m)}_j \neq I^{\psi}_j\right\}$.
  \item Repeat Steps 1--2 $R$ times to compute the average mis-selection rate for each candidate value, and select the one with the smallest average mis-selection rate.
\end{enumerate}
Depending on practical preferences, the mis-selection rate can be replaced by another criterion, such as the Matthews correlation coefficient. Thus, the learned posterior inclusion probabilities can be reused across different decision criteria, separating probability estimation from metric-specific variable selection without retraining the model. 

\subsection{Transformer for variable selection}\label{sec:transformer-vs}

Motivated by Proposition~\ref{prop:objective}, we consider a large function class represented by a Transformer; however, the standard Transformer architecture described in Section~\ref{sec:transformer} cannot be applied directly for three reasons: (i) its input and output shapes are incompatible with our problem; (ii) the response should not be treated as equivalent to a covariate; and (iii) it does not explicitly enforce the two permutation properties required for variable selection: invariance to permutations of the observations and equivariance to permutations of the covariates. 

To address these issues, we redesign the Transformer architecture to include embedding, self-attention, and output modules. 

\textbf{Entry-wise embedding}. Using the embedding rule in Equation~\eqref{eq:embedding}, we construct $\bm{H}^{(0)}_{ij}=\bm{E} X_{ij}$ for $i=1, \ldots, n$ and $j = 1, \ldots, p$, and construct $\bm{H}^{(0)}_{i, p+1}=\bm{E} Y_{i} + e_Y$ for $i=1, \ldots, n$, where $e_Y$ is a learnable parameter used to distinguish the response from the covariates. 

\textbf{Bidirectional self-attention}. We design a bi-attention module that first performs row-axis attention and then column-axis attention. Intuitively, row-axis attention utilizes information from the entire observed column of each variable, while column-axis attention utilizes information across the covariates and the response. Specifically, under row-axis attention, for each variable $j \in \{1,\ldots,p+1\}$, we take the $n$ embeddings associated with that variable and obtain
\[
(\widetilde{\bm{H}}_{1j},\ldots,
\widetilde{\bm{H}}_{nj})
=\texttt{SA}_{r}(\bm{H}^{(0)}_{1j},\ldots,
\bm{H}^{(0)}_{nj}),
\]
where $\texttt{SA}_{r}$ is a self-attention module as described in Section~\ref{sec:transformer} and is applied to each variable. We then apply self-attention along the column axis; i.e., for $i \in \{1,\ldots,n\}$, we obtain
\[
(\bm{H}^{(1)}_{i1},\ldots,\bm{H}^{(1)}_{i,p+1})
=\texttt{SA}_{c}(\widetilde{\bm{H}}_{i1},\ldots,
\widetilde{\bm{H}}_{i,p+1}),
\]
where the self-attention module $\texttt{SA}_{c}$ is applied to each observation. Then, $\bm{H}^{(1)}$ goes through another bidirectional self-attention module to obtain $\bm{H}^{(2)}$. In total, we apply the bidirectional self-attention module $K$ times to obtain $\bm{H}^{(K)}$. 

\textbf{Output module}. This module produces a $p$-dimensional vector of inclusion probabilities. For each observation $i$ and covariate $j$, we first combine the covariate representation $\bm{H}^{(K)}_{ij}$ with the response representation $\bm{H}^{(K)}_{i,p+1}$ to form $\bm{h}_{ij}=\sigma\!\left(\bm{W}_R\bm{H}^{(K)}_{i,p+1}+\bm{W}_C\bm{H}^{(K)}_{ij}\right)\in\mathbb{R}^{m}$, where $\bm{W}_R,\bm{W}_C\in\mathbb{R}^{m\times m}$ are learnable matrices. We then aggregate the information across observations by
\begin{align*}
\bm{z}_j =\sum_{i=1}^{n} \frac{\exp\!\left(\bm{q}^{\top}\bm{h}_{ij}/\sqrt{m}\right)}
{\sum_{k=1}^{n}\exp\!\left(\bm{q}^{\top}\bm{h}_{kj}/\sqrt{m}\right)}\,\bm{h}_{ij}\in\mathbb{R}^{m},
\end{align*}
where $\bm{q}\in\mathbb{R}^{m}$ is a learnable query vector. Stacking these vectors gives the matrix $\bm{Z}=(\bm{z}_1,\ldots,\bm{z}_p)^{\top}$, which is fed into an MLP to output the $p$-dimensional vector of inclusion probabilities $\sig\!\left(\texttt{MLP}(\bm{Z})\bm{w}_U+b_U\bm{1}_p\right) \in(0,1)^p$, where $\bm{w}_U\in\mathbb{R}^{m}$ and $b_U\in\mathbb{R}$ are learnable parameters.

\section{Simulation}\label{sec:simulation}

In this section, we conduct simulation studies to investigate the empirical performance of the proposed method, which we call MambaPFN, short for \underline{m}odel-\underline{a}gnostic \underline{M}arkov \underline{b}l\underline{a}nket selection with \underline{PFN}. In Section~\ref{sec:simu-selection}, we evaluate the quality of variable selection; in Section~\ref{sec:simu-scaling-raw}, we examine how the performance scales with the amount of pretraining data and the number of parameters; and in Section~\ref{sec:simu-computation}, we analyze the computational cost. 

\subsection{Variable selection evaluation}\label{sec:simu-selection}

We compare MambaPFN with several well-known variable selection methods, including Lasso \citep{tibshirani1996regression}, SCAD \citep{fan2001variable}, ABESS \citep{zhu2020polynomial}, DFS \citep{chen2021nonlinear}, STG \citep{yamada2020feature}, HSIC-Lasso \citep{yamada2014high}, and Knockoffs \citep{spector2022powerful}. These baselines cover both linear-model-based and nonlinear variable selection methods. We also compare with an amortized method for causal graph identification, from whose learned graph we extract the Markov blanket of the response. Specifically, we use CauScale \citep{peng2026causcale}, a state-of-the-art method for this task.

We report the F1-score of the selected subset and the area under the ROC curve (AUC), which evaluates the complete ranking of active versus inactive covariates; details of the evaluation of each method are described in Appendix~\ref{sec:experiment-details}. We evaluate selection accuracy on held-out datasets generated by Algorithm~\ref{alg:data-generation} (Section~\ref{sec:method-data}) and consider two cases: in the first, the number of observations varies over 100, 200, 400, 800, and 1600 with $p=20$; in the second, the number of observations is fixed at 800 while $p$ grows from 15 to 35.

\begin{figure}[t]
  \includegraphics[width=\linewidth]{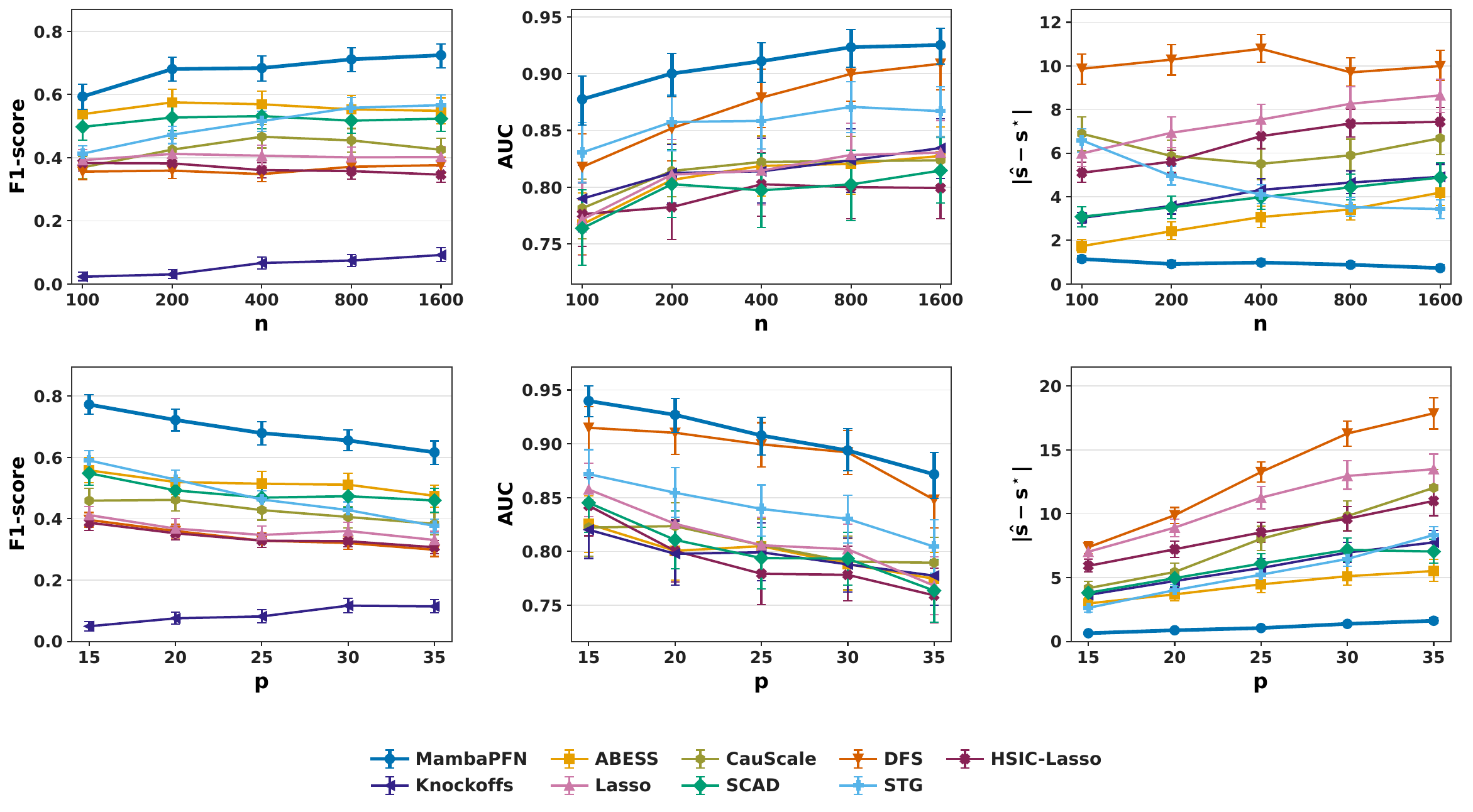}
  \caption{Comparison of baselines and MambaPFN on variable selection. Each row contains three panels showing the F1-score, AUC, and absolute model-size gap, respectively. The upper row varies $n\in\{100,200,400,800,1600\}$ with $p=20$, whereas the lower row varies $p\in\{15,20,25,30,35\}$ with $n=800$.}\label{fig:in-distribution-selection}
\end{figure}

The numerical results are summarized in Figure~\ref{fig:in-distribution-selection}. As the sample size increases, the AUC and F1-score of the methods generally increase, as expected. MambaPFN typically attains the highest F1-score and AUC, reflecting its effectiveness. Moreover, when the sample size is large, the AUC of MambaPFN is high, demonstrating that it separates active variables from irrelevant ones well. STG and DFS also perform reasonably well because both capture nonlinearity when selecting variables, which allows them to surpass linear-model-based methods such as SCAD and Lasso. It is also worth comparing with CauScale, which MambaPFN outperforms, probably because learning the full causal graph is more difficult. Finally, the rightmost panels of Figure~\ref{fig:in-distribution-selection} show that MambaPFN maintains a substantially smaller absolute model-size gap than the baselines, and this gap approaches zero as the sample size increases. 

\subsection{Scaling law}\label{sec:simu-scaling-raw}

MambaPFN is pretrained once and then applied without parameter updates. Its selection accuracy is therefore determined at the pretraining stage, and in particular by two resources: the number of pretraining datasets $M$ and the number of network parameters $N$. In this section, we examine whether the performance of MambaPFN improves predictably as $M$ and $N$ increase, in the spirit of the scaling laws reported for language models \citep{kaplan2020scaling}. To separate the two resources, we follow \citet{kaplan2020scaling} and vary one of them while holding the other fixed. 

We generate the pretraining data in advance. Specifically, we draw a pool of $3.2\times 10^{5}$ datasets. From this pool, we form six \textit{nested} training sets consisting of the first $9.6\times10^{3}$, $1.92\times10^{4}$, $4\times10^{4}$, $8\times10^{4}$, $1.6\times10^{5}$, and $3.2\times10^{5}$ datasets. As for model size, following \citet{kaplan2020scaling}, we increase the width and the depth jointly and otherwise keep the architecture. The numbers of parameters range from $5.9\times10^{4}$ to $1.1\times10^{7}$. When one factor varies, the other is fixed at its largest value. We evaluate the validation loss at $25$ checkpoints equally spaced on a logarithmic scale during training, and report the average of its three smallest values over these checkpoints\footnote{Selecting the best checkpoints plays the role of early stopping when a small training set is revisited many times, while averaging three of them reduces the fluctuation of a single checkpoint.}. The validation set is the fixed set of $2{,}048$ held-out DGPs, generated from a seed stream disjoint from the pretraining pool, following \citet{li2025mars}. 

\begin{figure}[t]
\centering
\includegraphics[width=1.0\linewidth]{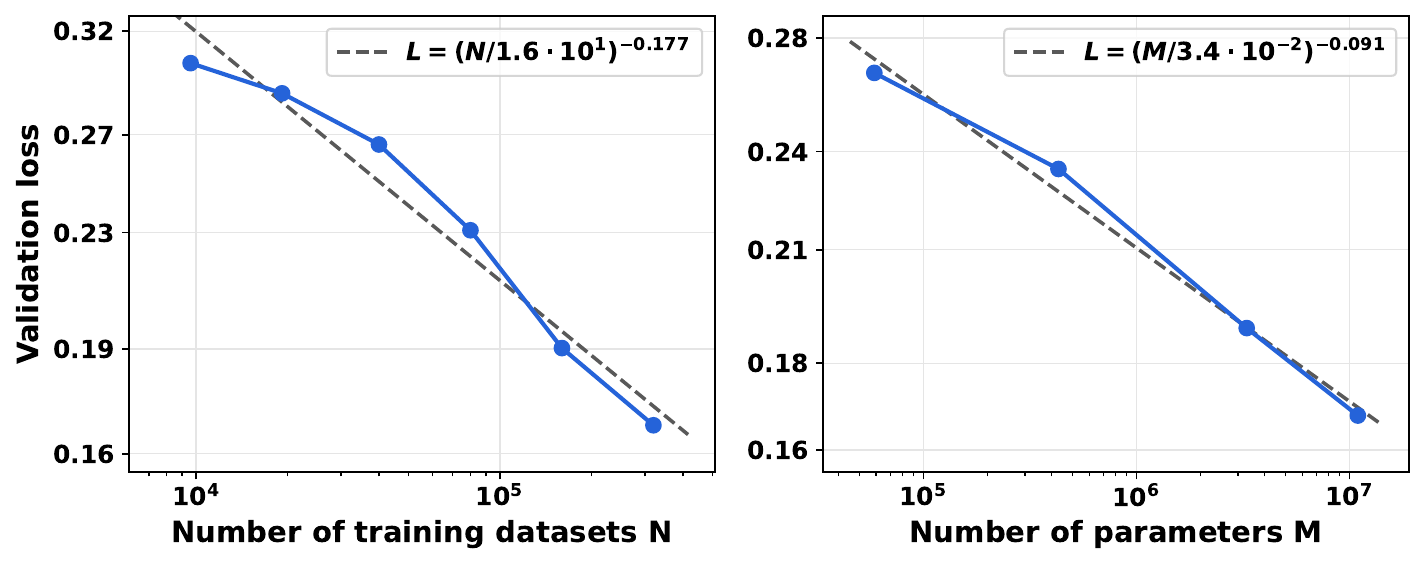}
\caption{Validation log-loss of MambaPFN on logarithmic axes. Left: the largest model trained on nested training sets of $M$ datasets. Right: models with $N$ parameters trained on the full pool ($M=3.2\times10^{5}$). Each point is the average of the three smallest validation losses of one run over $25$ checkpoints, and the lines are the fitted power laws.}\label{fig:scaling-raw}
\end{figure}

The results are summarized in Figure~\ref{fig:scaling-raw}. The validation loss decreases with both $N$ and $M$ over the ranges considered. These trends therefore indicate that the estimated inclusion probabilities move closer to the posterior inclusion probabilities as $N$ and $M$ grow, and suggest that the accuracy of MambaPFN can be further improved by scaling up the pretraining. 

\subsection{Computational analysis}\label{sec:simu-computation}

\begin{figure}[t]
\includegraphics[width=\linewidth]{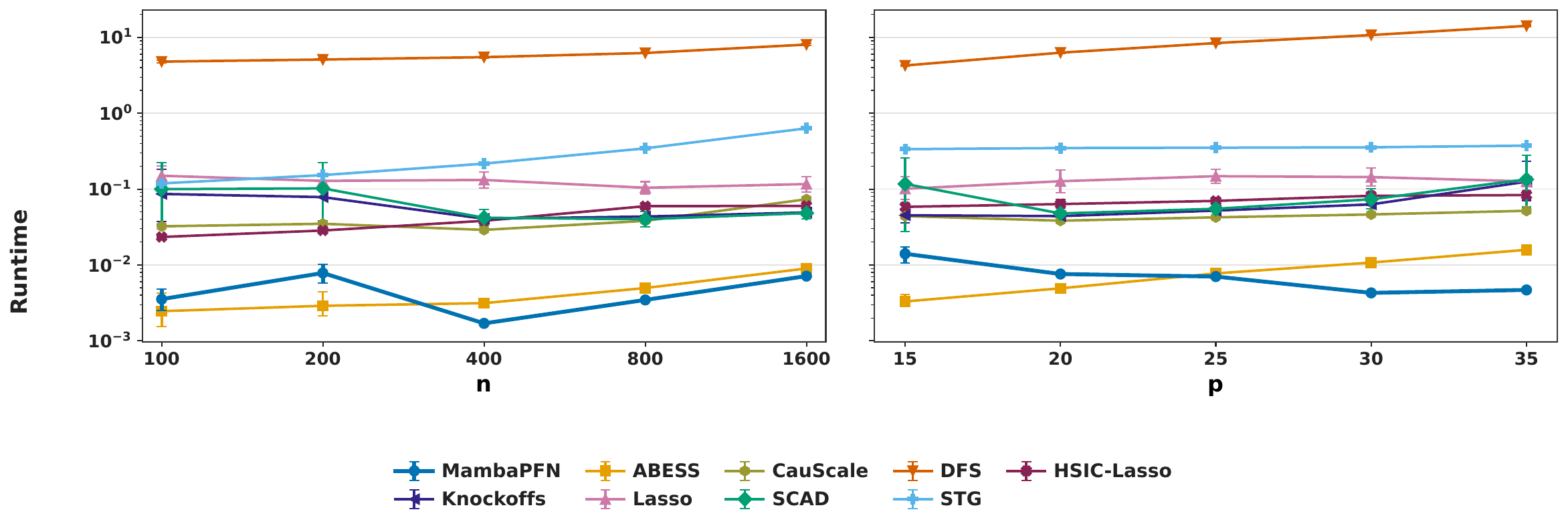}
\caption{Runtime comparison of MambaPFN with existing methods. The left panel varies $n$ with $p=20$, whereas the right panel varies $p$ with $n=800$.}\label{fig:runtime}
\end{figure}

In this section, we compare the computational cost of MambaPFN with that of the baselines. We consider two cases: in the first, we fix the dimension and increase the sample size; in the second, we fix the sample size and increase the dimension. We measure runtime in seconds, and the results are summarized in Figure~\ref{fig:runtime}. 

MambaPFN and ABESS have the shortest runtimes, and MambaPFN is faster than Lasso, even though it tackles a more challenging, model-agnostic setting. It is nearly 10 times faster than model-agnostic methods such as Knockoffs and HSIC-Lasso, and it is also faster than SCAD, which relies on a linear model. It is also about 10 times faster than CauScale, which spends more time recovering the full graph. These results confirm the computational merits of MambaPFN for selecting variables with nonlinear effects.

\section{Facebook Comments Data Analysis}\label{sec:real-data}

In this section, we evaluate MambaPFN on a real-world dataset for predicting the volume of comments on Facebook posts. Social media platforms such as Facebook seek to maximize the value of promoting a post. If a post shows a strong growth trend, promoting it may increase its traffic and generate additional revenue. At the same time, rapidly growing posts may carry a higher risk of spreading misinformation or provoking heated discussion; identifying such posts allows platforms to allocate human resources for timely monitoring and moderation.

To gain further insight into such growth trends, \citet{sing2015comment} collected a Facebook comment-volume dataset that is publicly available from the UCI repository\footnote{\url{https://archive.ics.uci.edu/dataset/363/facebook+comment+volume+dataset}}. The response of interest is the number of new comments received by a post. The dataset contains 53 covariates: 4 post-related variables, such as the number of likes; 25 variables comprising five summary statistics of historical comment counts (minimum, maximum, mean, median, and standard deviation); and 24 variables related to the timing and characteristics of the post (e.g., day-of-week indicators and the number of times the post was shared). Because this real-world dataset was not used to train the Transformer, it represents a challenging out-of-distribution setting. The dataset contains 40,949 observations.

We apply MambaPFN to the Facebook dataset to investigate which variables are most important for predicting the response. For comparison, we also consider the baselines introduced in Section~\ref{sec:simulation}\footnote{DFS is excluded because variable selection takes more than 10 hours on the same computing device.}. Because the relevant variables are unknown in this real-world dataset, we use predictive performance on the test set as a surrogate criterion. Specifically, we randomly allocate 30\% of the observations to variable selection, 30\% to fitting an MLP using only the selected variables, and 40\% to evaluating its predictive performance. We repeat the complete split-and-evaluate procedure 10 times to reduce split randomness. We also include a setting in which all variables are selected.

\begin{figure}[t]
  \includegraphics[width=\linewidth]{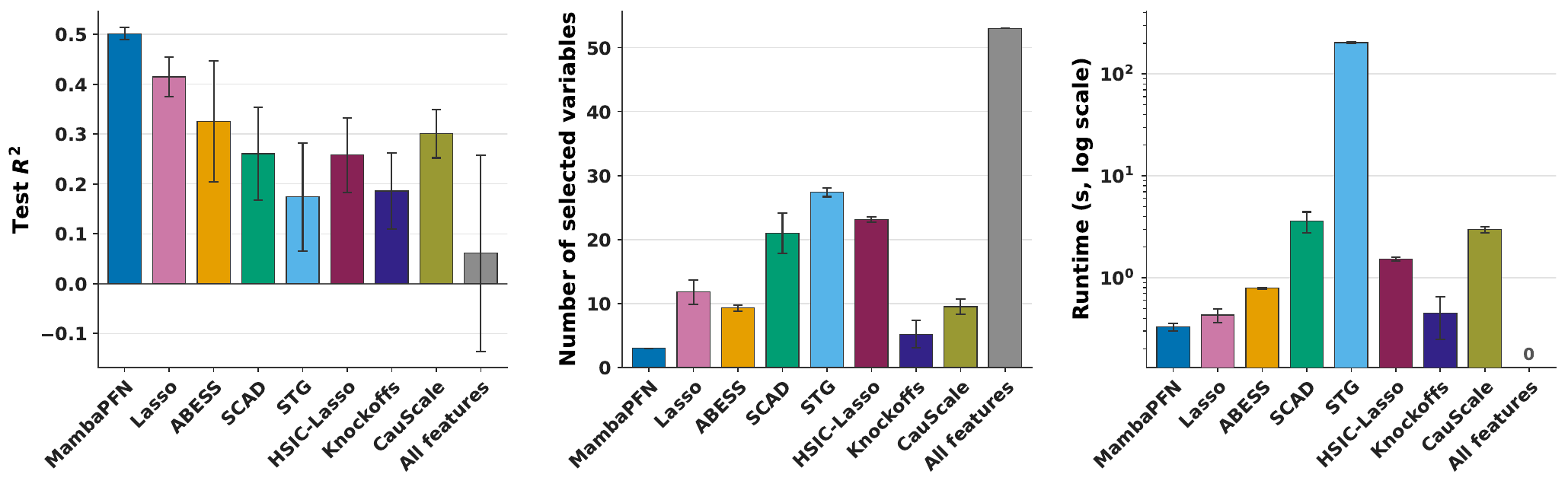}
  \caption{Comparison of methods on the Facebook comment dataset. The left panel shows the test-set $R^2$, the middle panel shows the number of selected variables, and the right panel shows the runtime of each method. Error bars show one standard error across the 10 repeated data splits.}
  \label{fig:real-data}
\end{figure}
The results are reported in Figure~\ref{fig:real-data}. MambaPFN achieves the highest $R^2$, suggesting that its selected variables have a strong association with the response. In contrast, naively fitting an MLP using all variables yields a substantially lower test $R^2$ than performing variable selection, reflecting the advantage of variable selection. MambaPFN also significantly outperforms all other methods and the improvement upon the best baseline approaches 20\%. Meanwhile, it selects the fewest variables, suggesting that it effectively captures nonlinear dependence with a more parsimonious subset. Finally, it has the shortest runtime, taking less than one second to analyze the dataset, which is about 5 and 9 times faster than HSIC-Lasso and CauScale, the two model-agnostic methods with the most competitive predictive performance. 

\begin{figure}[t]
  \includegraphics[width=\linewidth]{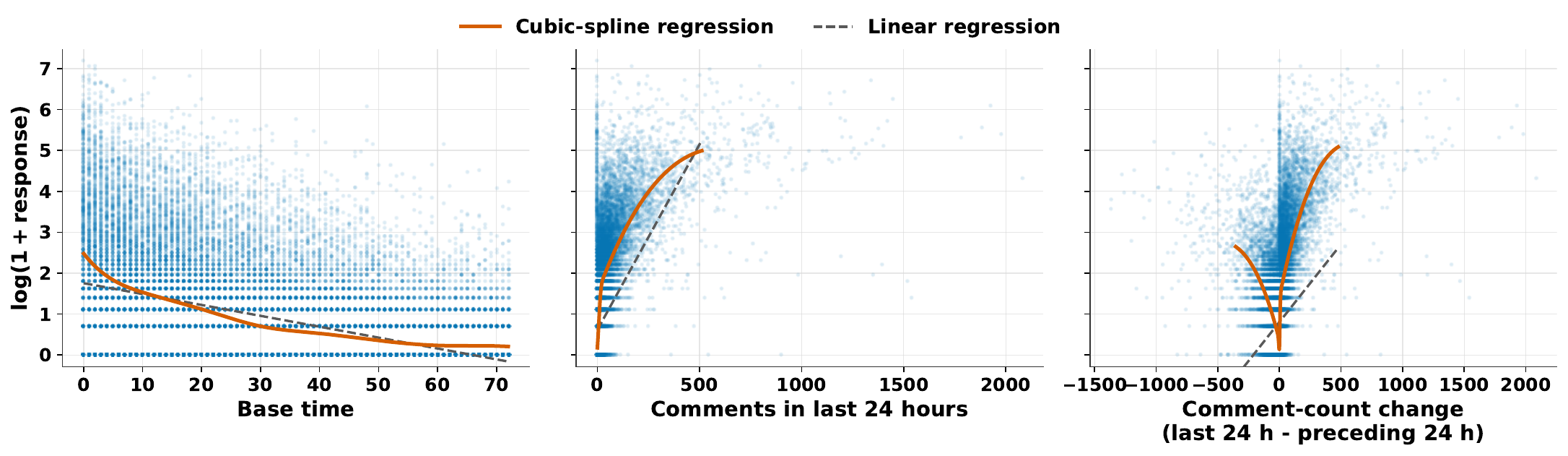}
  \caption{Scatterplots of the three selected variables against the $\log$-transformed response. Each panel shows one selected variable. The red curve denotes the nonlinear regression fit, while the black dashed line denotes the linear regression fit. The estimated posterior probabilities of the three variables are 0.94, 0.89, and 0.71, respectively.}
  \label{fig:real-data-vars}
\end{figure}
Finally, it is worth noting that MambaPFN selects three variables: (i) the time elapsed between publication of the post and the current prediction time (denoted as Base time); (ii) the number of comments in the last 24 hours; and (iii) the temporal difference in comment counts, measured over 24-hour intervals. Their associations with the response are visualized in Figure~\ref{fig:real-data-vars}, and all three exhibit some degree of nonlinearity. The last variable shows the strongest nonlinearity, with a roughly V-shaped association: a large positive value indicates accelerating comment activity, whereas a large negative value can correspond to a previously active post that is now cooling. Values near zero largely represent inactive posts, so this variable captures both the direction and the scale of comment activity; this may explain why linear-model-based methods require additional variables to represent the same pattern. The number of comments in the last 24 hours has a strong positive but nonlinear association with the response, making it a direct proxy for the post's current popularity. In contrast, Base time is negatively associated with future comments, consistent with an attention-decay process that is steep shortly after publication and then gradually levels off.

\section{Conclusion and Discussion}\label{sec:conclusion}

This paper brings the viewpoint of the bitter lesson to variable selection, a long-standing problem in statistics, and develops an amortized variable selection method, MambaPFN. We provide a principled implementation of this method through the design of the pretraining data, the loss, and the architecture. Both synthetic and real-world data analyses confirm the empirical merits of this method.  

Several directions deserve further investigation. First, MambaPFN currently returns only the variable selection results, and a separate MLP is trained for prediction, forming a two-step procedure. It would be interesting to extend the current architecture to make sample-wise predictions based on the posterior inclusion probabilities, yielding an interpretable machine learning method that outputs both the selected variables and predictions based on them. Second, although MambaPFN saves considerable time at the selection stage, pretraining the Transformer is computationally costly. It would be interesting to consider other architectures, such as selective state space models \citep{gu2023mamba} or grouped-query attention \citep{ainslie2023gqa}, to speed up training and inference. 

\bibliography{bibliography}

\newpage
\phantomsection\label{supplementary-material}
\appendix

\section{Proofs}
\label{app:theory-details}
\subsection{Proof of Proposition~\ref{prop:coverage}}
\label{app:coverage}

\begin{proof}
Choose $\psi\in\mathcal{G}$ such that $Y=V_{p+1}$ has no children and its parents are the active covariates. Equation~\eqref{eq:structure-model} then gives
\[
Y=f\!\left(X_{\operatorname{pa}_\psi(p+1)}\right)+\varepsilon_{p+1},
\]
where $\varepsilon_{p+1}$ is independent of $X$ because no covariate is generated from $Y$.

For a generalized additive model, take
\[
f(u)=\beta_0+\sum_{j\in\operatorname{pa}_\psi(p+1)}g_j(u_j)
\]
in $\mathcal{F}$. Substitution yields $Y=\beta_0+\sum_j g_j(X_j)+\varepsilon_{p+1}$, with $g_j=0$ outside $\operatorname{pa}_\psi(p+1)$; choosing $g_j(u)=\beta_j u$ gives the linear model as a special case. For an interaction model, take
\[
f(u)=\beta_0+\sum_j\beta_j u_j+\sum_{j<k}\gamma_{jk}u_ju_k,
\]
and choose $\operatorname{pa}_\psi(p+1)$ to contain exactly the covariates having a nonzero main or interaction term. Equation~\eqref{eq:structure-model} then gives the corresponding interaction model. Thus both model classes are included in the structural model.
\end{proof}

\subsection{Proof of Lemma~\ref{lemma:graph-design}}
\label{app:graph-design}

\begin{proof}
Strict positivity of the factors in Equation~\eqref{eq:prob-factorize} implies $S^\star\subseteq\textup{MB}_\psi(Y)$ \citep{pearl2014probabilistic,koller2009probabilistic}. For the reverse inclusion, fix $X_j\in\textup{MB}_\psi(Y)$. If $Y\not\perp X_j\mid X_{-j}$, then $j\in S^\star$: otherwise, weak union applied to the defining conditional independence for $S^\star$ would give $Y\perp X_j\mid X_{-j}$.

Only the factor of $Y$ and those of its children remain in the conditional density of $Y$ given $X$. Assumption~\ref{con:generate-graph} ensures that $X_j$ occurs in exactly one of these factors: it is either a parent of $Y$, a child of $Y$, or another parent of one child of $Y$. We use two elementary properties of the noise density $g$. First, $g(t-a)=\kappa g(t-b)$ for almost every $t$ implies $a=b$: integration gives $\kappa=1$, and a nonzero translation would make an integrable density periodic. Second, log-concavity implies
\[
g(t-a)g(t-d)=g(t-b)g(t-c)\ \text{ for almost every }t
\quad\Longrightarrow\quad
\{a,d\}=\{b,c\}.
\]
Indeed, the Hellinger autocorrelation $u\mapsto\int\sqrt{g(t)g(t+u)}\,dt$ is even and log-concave, and is strictly decreasing in $|u|$ by the first property and the Cauchy--Schwarz inequality \citep{prekopa1973logarithmic}. Integrating the square root of the displayed identity gives equal separations between the two pairs of shifts; after aligning their signs, unequal midpoints would make the corresponding positive, integrable centered product periodic. Hence the two pairs coincide.

Suppose, contrary to the claim, that $Y\perp X_j\mid X_{-j}$. If $X_j$ is a parent of $Y$, the factor associated with $Y$ shows that two translates of its noise density must be proportional as $x_j$ varies. The first property forces the corresponding values of $f$ to be equal, contradicting Assumption~\ref{con:generate-function}. The same argument, with $y$ and $x_j$ interchanged, applies when $X_j$ is a child of $Y$.

Finally, suppose that $X_j$ is another parent of a child of $Y$. Conditional independence and the factor associated with that child give, for almost every $(y,y',x_j,x_j')$, the second displayed identity with the four shifts
\[
f(y,x_j,\ldots),\quad f(y',x_j,\ldots),\quad
f(y,x_j',\ldots),\quad f(y',x_j',\ldots).
\]
The resulting equality of the two multisets requires either both $y$-sections or both $x_j$-sections to agree. Fixing a nonconstant $x_j$-section forces the $y$-sections to be constant on a set of positive measure, while fixing a nonconstant $y$-section forces the $x_j$-sections to be constant. Either conclusion contradicts Assumption~\ref{con:generate-function}. Thus $Y\not\perp X_j\mid X_{-j}$ in all three cases, so $\textup{MB}_\psi(Y)\subseteq S^\star$. Combining the two inclusions proves $\textup{MB}_\psi(Y)=S^\star$.
\end{proof}

\subsection{Proof of Proposition~\ref{prop:objective}}
\begin{proof}
Conditioning on $\mathcal{D}_n$ in Equation~\eqref{eq:sel-loss} gives
\begin{align*}
\mathcal{R}(\theta)
=-\frac{1}{p}\sum_{j=1}^p\mathbb{E}_{\mathcal{D}_n}
\Big[&\mathbb{P}_{\psi\sim\mathcal{P}}\!\left(I_j^\psi=1\mid\mathcal{D}_n\right)
\log T_{\theta,j}(\mathcal{D}_n)\\
&+\mathbb{P}_{\psi\sim\mathcal{P}}\!\left(I_j^\psi=0\mid\mathcal{D}_n\right)
\log\!\left\{1-T_{\theta,j}(\mathcal{D}_n)\right\}\Big].
\end{align*}
Substituting the assumed expression for $T_{\bar\theta,j}(\mathcal{D}_n)$ and subtracting yields
\begin{align*}
\mathcal{R}(\theta)-\mathcal{R}(\bar\theta)
=\frac{1}{p}\sum_{j=1}^p\mathbb{E}_{\mathcal{D}_n}
\operatorname{KL}\!\left(
\operatorname{Bern}\!\left\{\mathbb{P}_{\psi\sim\mathcal{P}}(I_j^\psi=1\mid\mathcal{D}_n)\right\}
\,\middle\Vert\,
\operatorname{Bern}\!\left\{T_{\theta,j}(\mathcal{D}_n)\right\}
\right)\geq0.
\end{align*}
Hence $\mathcal{R}(\theta)\geq\mathcal{R}(\bar\theta)$ for every $\theta$.
\end{proof}

\subsection{Proof of Proposition~\ref{prop:posterior-limit}}
\begin{proof}
To compare different context lengths on the same data, we take the contexts to be nested: $Z_1,Z_2,\ldots$ is an infinite sequence of observations with $Z_i=(X_i,Y_i)$, and $\mathcal{D}_n=\{Z_1,\ldots,Z_n\}$ consists of its first $n$ terms, so that $\mathcal{D}_n\subseteq\mathcal{D}_{n+1}$.

Let $Z_{1:\infty}=(Z_1,Z_2,\ldots)$ denote the whole sequence. Throughout the proof, $\mathbb{P}_{\psi\sim\mathcal{P}}$ and $\mathbb{E}_{\psi\sim\mathcal{P}}$ refer to the joint law under which $P_{\psi,f}$ is drawn from $\mathcal{P}$ and, given $P_{\psi,f}$, $Z_1,Z_2,\ldots$ are independent with law $P_{\psi,f}$. Write
\[
\eta_j(\mathcal{D}_n)=\mathbb{P}_{\psi\sim\mathcal{P}}\left(I^{\psi}_j=1\mid\mathcal{D}_n\right)=\mathbb{E}_{\psi\sim\mathcal{P}}\left(I^{\psi}_j\mid Z_1,\ldots,Z_n\right),
\]
so that $T_{\bar\theta,j}(\mathcal{D}_n)=\eta_j(\mathcal{D}_n)$ by the definition of $\bar\theta$ in Proposition~\ref{prop:objective}. Since $I^{\psi}_j$ is bounded, $\{\eta_j(\mathcal{D}_n)\}_{n\geq1}$ is a martingale, and L\'evy's upward theorem \citep[Section~14.2]{williams1991} gives
\[
\eta_j(\mathcal{D}_n)\to\mathbb{E}_{\psi\sim\mathcal{P}}\left(I^{\psi}_j\mid Z_{1:\infty}\right)
\qquad\text{almost surely and in }L_1 .
\]
It remains to show that $I^{\psi}_j$ is almost surely a function of $Z_{1:\infty}$, in which case $\mathbb{E}_{\psi\sim\mathcal{P}}(I^{\psi}_j\mid Z_{1:\infty})=I^{\psi}_j$.

Let $F_{\psi,f}$ be the distribution function of $P_{\psi,f}$ on $\mathbb{R}^{p+1}$ and $\widehat F_n$ the empirical distribution function of $\mathcal{D}_n$. Conditionally on $P_{\psi,f}$, the observations are independent with law $P_{\psi,f}$, and the strong law of large numbers gives $\widehat F_n(t)\to F_{\psi,f}(t)$ almost surely for each $t\in\mathbb{Q}^{p+1}$; since $\mathbb{Q}^{p+1}$ is countable, the convergence holds simultaneously for all such $t$, and integrating over $P_{\psi,f}\sim\mathcal{P}$, it holds with probability one. A distribution function is determined by its values on $\mathbb{Q}^{p+1}$ by right-continuity, and it determines the distribution, so $P_{\psi,f}$ is almost surely a function of $Z_{1:\infty}$. By Lemma~\ref{lemma:graph-design}, $I^{\psi}_j=I_j=\mathbb{I}(j\in S^\star)$, where $S^\star$ is the minimal Markov blanket of $Y$ under $P_{\psi,f}$ (Section~\ref{sec:preliminaries}); hence $I^{\psi}_j$ is determined by $P_{\psi,f}$ and is therefore almost surely a function of $Z_{1:\infty}$. Consequently, $\mathbb{E}_{\psi\sim\mathcal{P}}(I^{\psi}_j\mid Z_{1:\infty})=I^{\psi}_j$ almost surely, and $T_{\bar\theta,j}(\mathcal{D}_n)\to I^{\psi}_j$ almost surely.
\end{proof}

\section{Experimental details}\label{sec:experiment-details}
\subsection{Details of Section~\ref{sec:simu-selection}}
\textbf{Experimental design}. We evaluate all methods on held-out data-generating processes (DGPs) independently generated by Algorithm~\ref{alg:data-generation} (Section~\ref{sec:method-data}). In the sample-size experiment, we fix $p=20$ and use $n\in\{100,200,400,800,1600\}$; each point is averaged over 200 DGP replicates. Within a replicate, the structural parameters, response variable, and active set are shared across context lengths, and all methods receive the same dataset. Covariates are randomly permuted before evaluation. For the classical baselines, both the covariates and response are standardized within each dataset; MambaPFN applies its own internal standardization.

\textbf{Implementation and selection rules}. Each method determines its own model size, and the true support size is used only for evaluation. In particular,
\begin{itemize}[leftmargin=*]
  \item Lasso uses five-fold cross-validation and selects the nonzero coefficients. ABESS uses the support selected by its official implementation under the extended special information criterion, while SCAD selects the minimum-BIC point along its regularization path. The functions/libraries for Lasso, ABESS and SCAD are \texttt{LassoCV}, \texttt{abess} \citep{zhu2022abess}, and \texttt{picasso} \citep{ge2019picasso}, respectively. 
  \item DFS fits candidate support sizes from $0$ to $p$ and selects the minimum-BIC candidate. Our PyTorch implementation follows the authors' reference code\footnote{\url{https://github.com/cyustcer/Deep-Feature-Selection}} for \citet{chen2021nonlinear}. STG selects variables whose learned gate probabilities exceed $0.5$.
  \item HSIC-Lasso applies three-fold positive \texttt{sklearn.linear\_model.LassoCV} to the block-HSIC design; for $n>800$, it uses a random subsample of 800 observations. A reference implementation is provided by the \texttt{pyHSICLasso} library\footnote{\url{https://github.com/riken-aip/pyHSICLasso}}. Knockoffs is implemented with the \texttt{knockpy} library\footnote{\url{https://github.com/amspector100/knockpy}} and uses Gaussian model-X knockoffs with the minimum-variance-reconstructability construction, a Lasso feature statistic, Ledoit--Wolf covariance shrinkage, and target FDR $0.1$.
  \item CauScale uses the authors' official implementation\footnote{\url{https://github.com/OpenCausaLab/CauScale}} and the released synthetic-data checkpoint\footnote{\url{https://huggingface.co/OpenCausaLab/causcale-model}} to estimate a directed adjacency matrix for $(Y,X)$. After thresholding edge probabilities at $0.5$, we select the estimated Markov blanket of $Y$, consisting of its parents, children, and other parents of its children.
\end{itemize}
Thus, no method is given $|S^\star|$ when constructing its selected set. Continuous scores used for AUC are also method-specific: these are inclusion probabilities for MambaPFN, path-entry or coefficient scores for the penalized methods, learned gate scores for DFS and STG, the knockoff statistic for Knockoffs, and Markov-blanket edge scores for CauScale.

\textbf{Evaluation criteria}. Let $\widehat S$ and $S^\star$ denote the selected and true active sets. Precision and recall are defined as
\[
  \operatorname{Precision}=\frac{\operatorname{TP}}
  {\operatorname{TP}+\operatorname{FP}},
  \qquad
  \operatorname{Recall}=\frac{\operatorname{TP}}
  {\operatorname{TP}+\operatorname{FN}}.
\]
Precision is the proportion of selected variables that are truly active, whereas recall is the proportion of truly active variables recovered by the method. Their harmonic mean gives
\[
  \operatorname{F1}
  =\frac{2\,\operatorname{Precision}\,\operatorname{Recall}}
  {\operatorname{Precision}+\operatorname{Recall}}.
\]
When no variable is selected, precision and the F1-score are set to zero. The ROC-AUC is computed from the continuous variable scores within each dataset and equals the proportion of active--inactive pairs for which the active variable receives the larger score, with ties assigned weight $1/2$. We also report the absolute model-size gap $\lvert |\widehat S|-|S^\star|\rvert$ to distinguish ranking errors from systematic over- or under-selection. Reported curves are replicate means, and their error bars are pointwise 95\% intervals obtained from 2,000 bootstrap resamples of complete DGP replicates; for the sample-size experiment, this preserves the pairing across context lengths. Runtime, when reported, is wall-clock time per dataset for method-specific fitting and scoring and excludes DGP generation, metric calculation, and plotting. MambaPFN and CauScale inference was performed on a single NVIDIA A100 GPU with 80~GB of memory, while the remaining baselines ran on the allocated CPU cores.

\subsection{Details of Section~\ref{sec:simu-scaling-raw}}

All runs share the same number of gradient steps and the same optimization settings, so a run cycles over its training set several times, e.g., four times for the full pool\footnote{We limit the repetition on the full pool to four passes because, for language models, up to four passes over the same data have been found to yield a loss close to that obtained with the same amount of fresh data \citep{muennighoff2023scaling}.}. Fixing the number of steps ensures that a larger $M$ brings more distinct data rather than more optimization.

When studying the effect of model size, we fix the training set at the full pool, $M=3.2\times10^{5}$, and vary the model size. As for studying the effect of pretraining data scale, we fix the largest model with $N=1.1\times10^{7}$, so that model capacity is unlikely to be the bottleneck, and vary $M$ over the six nested training sets. The two experiments share one run, namely the largest model trained on the full pool. 

\subsection{Details of Section~\ref{sec:real-data}}

\textbf{Dataset preparation}. We use Variant~1 of the Facebook Comment Volume training data from the UCI repository, containing 40,949 observations, 53 predictors, and the number of comments received in the following $H$ hours as the response. The fourth predictor, page category, is treated as categorical. The remaining predictors are treated as numeric, and the two downstream figures retain the original feature definitions supplied with the dataset.

\textbf{Leakage-separated evaluation}. For each of 10 seeded repetitions, we randomly divide the data into 12,284 observations for variable selection, 12,284 for downstream MLP fitting, and 16,379 for final testing; the remaining two observations produced by integer rounding are unused. The three subsets are mutually disjoint, and every selector receives the same selection subset within a repetition. Median imputation and standardization for feature selection are estimated using only the selection subset. The MambaPFN checkpoint was pretrained with $p\le 35$, so its application to the 53 Facebook predictors also evaluates dimension extrapolation.

\textbf{Downstream preprocessing and prediction}. After variable selection, numeric predictors are median-imputed and standardized using only the downstream training subset, while the page-category predictor is one-hot encoded using levels observed in that subset. The response is standardized using the downstream training subset but is not log-transformed; all test metrics are computed on the original response scale. The downstream predictor is an MLP with hidden widths 256 and 128, GELU activations, and dropout 0.1. It is optimized by AdamW with learning rate $10^{-3}$, weight decay $10^{-4}$, and batch size 2,048 for at most 100 epochs. Within the downstream training subset, 15\% of the observations are used for early stopping with patience 10; the selected number of epochs is then used to refit the MLP on the complete downstream training subset. Test $R^2$ is evaluated once per repetition, and Figure~\ref{fig:real-data} reports means and standard errors across the 10 repetitions. The all-variable MLP follows the same preprocessing and fitting protocol.

\textbf{Full-data descriptive analysis}. To interpret the variables selected by MambaPFN, we separately apply it to all 40,949 observations after full-data median imputation and standardization. The observations are partitioned into balanced, nonoverlapping blocks of at most 1,000 rows, the raw logits are averaged across blocks, and variables with inclusion probability above $0.5$ are selected. Figure~\ref{fig:real-data-vars} plots every observation using $\log(1+Y)$ and overlays a linear fit and a cubic-spline ridge fit with eight quantile-spaced knots; the fitted curves are drawn over the central 99\% of each predictor. 

\end{document}